\documentclass{article} 
\usepackage{iclr2022_conference,times}

\usepackage{amsmath,amsfonts,bm}

\def\eqref#1{equation~\ref{#1}}

\def\1{\bm{1}}

\DeclareMathAlphabet{\mathsfit}{\encodingdefault}{\sfdefault}{m}{sl}
\SetMathAlphabet{\mathsfit}{bold}{\encodingdefault}{\sfdefault}{bx}{n}

\usepackage{hyperref}
\usepackage{url}
\usepackage{graphicx}
\usepackage{subcaption}
\newcommand{\repeatthanks}{\textsuperscript{\thefootnote}}

\title{Document Structure aware \\Relational Graph Convolutional Networks \\for Ontology Population}

\iclrfinalcopy
\author{Abhay M Shalghar\thanks{Equal Contribution}, Ayush Kumar\repeatthanks \& Shobha G\\
RV College of Engineering\\
Bengaluru, India \\
\texttt{\{abhayms.cs18,ayushkumar.cs18,shobhag\}@rvce.edu.in} \\
\AND
Balaji Ganesan \& Aswin Kannan \\
IBM Research \\
Bengaluru, India \\
\texttt{\{bganesa1,aswkanna\}@in.ibm.com} \\
\AND
Akshay Parekh \\
IIT Guwahati \\
Guwahati, India \\
\texttt{akshayparakh@iitg.ac.in}
}

\begin{document}

\maketitle

\begin{abstract}
Ontologies comprising of concepts, their attributes, and relationships are used in many knowledge based AI systems. While there have been efforts towards populating domain specific ontologies, we examine the role of document structure in learning ontological relationships between concepts in any document corpus. Inspired by ideas from hypernym discovery and explainability, our method performs about 15 points more accurate than a stand-alone R-GCN model for this task.
\end{abstract}

\section{Introduction}

Ontology induction (creating an ontology) and ontology population (populating ontology with instances of concepts and relations) are important tasks in knowledge based AI systems. While the focus in recent years has shifted towards automatic knowledge base population and individual tasks like entity recognition, entity classification, relation extraction among other things, there are infrequent advances in ontology related tasks.

Ontologies are usually created manually and tend to be domain specific i.e. meant for a particular industry. For example, there are large standard ontologies like Snomed for healthcare, and FIBO for finance. However there are also requirements for cross domain ontologies for applications in data protection, meta-data management and data discovery in Data Fabric.

Creating and maintaining ontologies and related knowledge graphs is a laborious process. There have been few efforts in recent years to automate ontology population. \cite{chen2018on2vec} introduced constraints on relations that should be part of ontologies. \cite{guan2019link} et al proposed a method for Link Prediction in n-ary relational data. \cite{shen2020taxoexpan} used a R-GCN model for the related task of taxonomy expansion. We continue with the recent trend to use R-GCN to predict relation type (link type prediction) between entities.

\begin{figure}[htb]
\centering
    \includegraphics[height=8cm, width=0.8\textwidth]{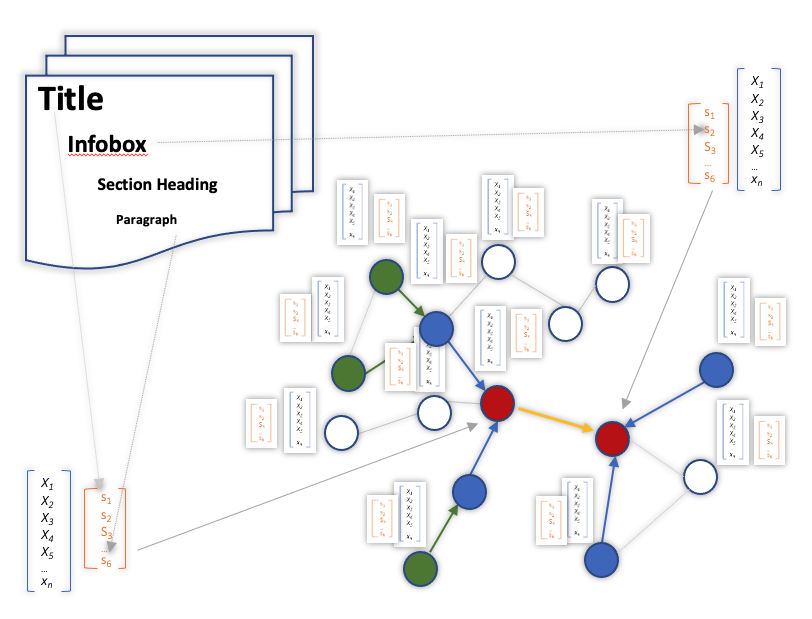}
    \caption{Document Structure can improve relation extraction using RGCN.}
    \label{fig:doc_structure}
\end{figure}

In this work, we focus on relation extraction using relational graph neural networks (RGCN) for ontology population. These ideas stem from observations of knowledge based systems in industrial applications. Many of these datasets used for this task are usually derived from formatted documents, but the document structure information is often discarded by converting to triples and other dataset formats. So for the relation extraction task, as shown in Figure \ref{fig:doc_structure} instead of starting from plain text sentences, we explore incorporating the \textit{document structure} information to improve accuracy in R-GCN models.

We summarize our contributions in this work as follows:
\begin{itemize}
    \item We propose a document structure measure (DSM) similar to other measures in hypernmy discovery to model ontological relationships between entities in documents.
    \item We experiment with different methods to incorporate the DSM vectors in state of the art relational graph convolutional networks and share our results.
\end{itemize}

\section{Related Work}

\cite{chen2018on2vec} incorporated hierarchy for relation extraction in Ontology Population. Our document structure measures differs from such works in a way that it not only capture hierarchy but are very general and encompass document summary, section titles, bulleted lists, highlighted text, infobox etc. We also extend this concept of document structure to personal data (binary).

\cite{nagpal2022fine} described a method to use rule based systems along with a neural model to improve fine grained entity classification. \cite{vannur2020data} described a method to augment the training data for better personal knowledge base population. \cite{ganesan2020link} uses graph neural networks to predict missing links between people in a property graph.

\cite{chiticariu2010systemt} presented a System T method of rule based information extraction. They have also shown a way to determine if the extraction information is relevant to a domain or otherwise \cite{wang2017towards}. \cite{madaan2017visual} has specifically dealt with extracting titles, section and subsection headings, paragraphs, and sentences from large documents. Additionally~\cite{agarwal2017cognitive} extracts structure and concepts from html documents in compliance specific applications.
 
\cite{zhang2018graph} proposed an extension of graph convolutional network that is tailored for relation extraction. Their model encodes the dependency structure over the input sentence with efficient graph convolution operations, then extracts entity-centric representations to make robust relation predictions. We derive some of our inspiration from the TaxoExpan work \cite{shen2020taxoexpan} that also uses RGCN to find hyponyms for taxonym expansion. We note that our work is however significantly more comprehensive in both the types of rules and number of such relationship extractions (not restricted to hypernym discovery). 
\section{Learning Document Structure}
\label{our_approach}

\begin{figure*}[htb]
    \includegraphics[width=\linewidth]{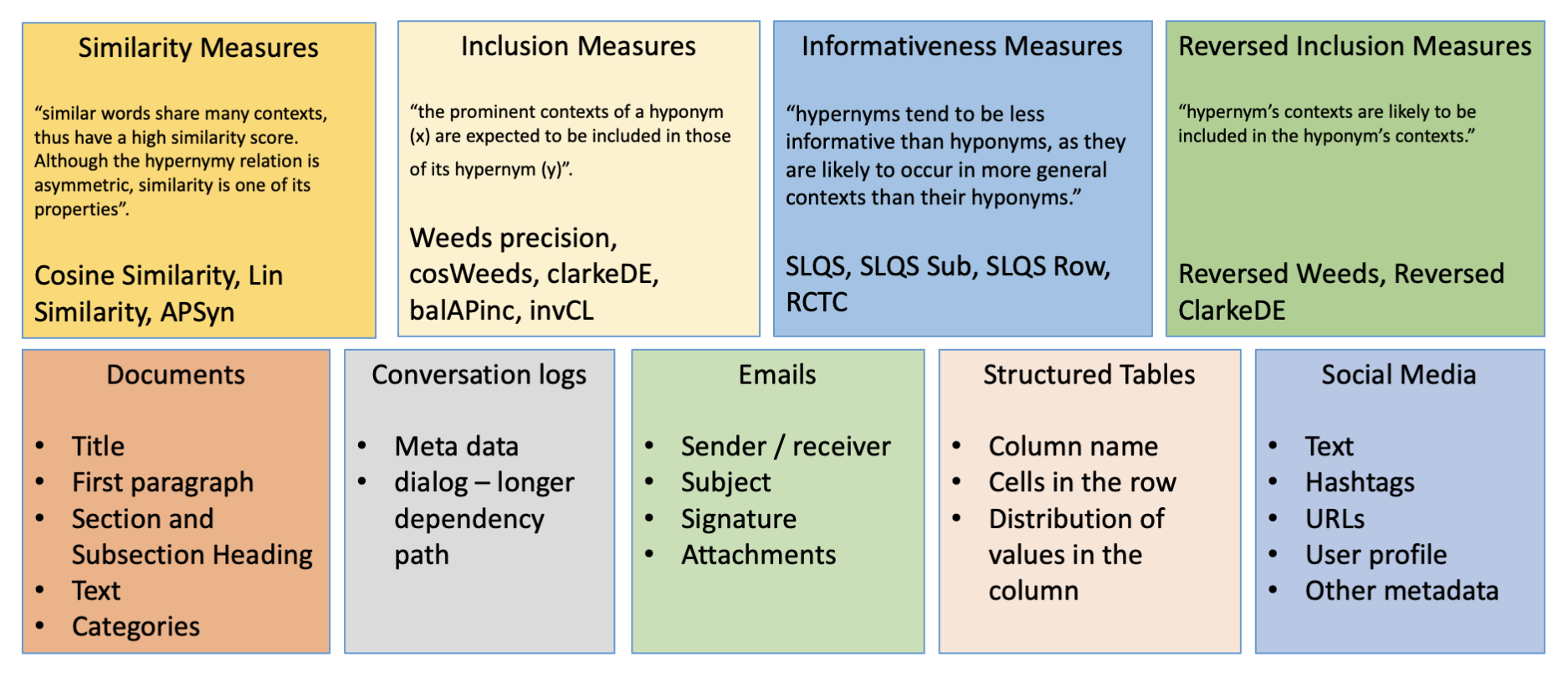}
    \caption{Document Structure in different document formats.}
    \label{fig:doc_structure_across_documents}
\end{figure*}

~\cite{chen2018on2vec} showed that some relations prove to be rich enough to go into the T-box (relations between Concepts), while other relations need only be present in the A-box. A corollary to the above classification of relations is the use of different measures to solve the hypernym discovery problem. These could be broadly classified as Similarity Measures (Cosine, Lin, ApSyn), Inclusion Measures  (ClarkeDE, balAPinc, invCL), Informativeness Measures (SLQS), and Reversed Inclusion Measures (Reversed ClarkeDE).

Inspired by these measures, we introduce a new measure based on document features and structure that helps identifying relationships for knowledge base population. We broadly term these under the umbrella of Document Structure Measures (DSM) and further deploy a Relational Graph Convolutional Network (RGCN) to learn the attributions of the same towards our larger goal. To validate the effectiveness of using DSM, we perform experiments on Wikipedia pages of entities in our datasets. Our focus on unstructured text and personal data in Wikipedia pages is primarily due to restrictions on using emails or social media posts for research. However, the approach is general to any type of document with a template. Some easy examples include  corporate documents, emails, tweets, logs, and chat transcripts.

\subsection{Document Structure Measure (DSM)}
\label{sec:measures}
To motivate the notational aspects, before defining our model, we shed some light on the broad term called ``document features''.
These can correspond to positioning of words, their hierarchy in a document, and even their formatting. We note that some of the features from embedding can also be cast into this framework, but we precisely mention that our structural features are generic and not tied to the aspects of any specific language. As an example, text within bullets can represent important data such as sensitive data, hypernyms, or even a short excerpt. As another isometric viewpoint, personal data like anonymized or pseudonymized credit card numbers can be presented in brackets, found in footnotes, and even given by fine worded text (small text) to mask readability. 

In our scope of work corresponding to ontologies, we choose ``personal information'' as our running example. These can say point to names, addresses, biographical details etc. and covers a very broad range of labels. As the most crucial aspect, we define the metric called importance probability, $\rho$. This measure signifies the factor by which a specific occurrence of text corresponds to a classification label in the space defined by document structure measure. Say, in our running example, if the text ``xxxx--56'' is found within brackets, we attribute a higher probability score $\rho$ to its classification as personal data. As another example in a slightly different context, say if the word ``red'' appears in bullets, and the word ``color'' appears in the main text, we attribute a higher probability score to the hypernym-hyponym relationship between the two words.

Next, we define vectors, $v_a(x)$, $v_b(x)$, and $v_c(x)$ to define the document structure measure corresponding to a word $x$. Here, $v_a(x)$ and $v_b(x)$ refer to relational measures and $v_c(x)$ refers to an absolute measure. More importantly, $v_a$ refers to a higher hierarchy and $v_b$ refers to a lower hierarchy. Say, in the example of ``red/color'' relationship above, let ``$x$'' denote ``red''. Then, $v_a(x)$ takes a lower value (say 0) and $v_b$ takes a higher value (say 1). Without loss of generality (this can take non-integral values also), we assume that 
the values taken by $v_a$, $v_b$, and $v_c$ are 0/1 (similar to an indicator function). Next, we further specify the fine details on the quantification of these vectors based on the logical document structure rules. Noting space considerations, we discuss in detail in about relational DSM vectors in the next subsection. Details regarding absolute DSM vectors can be found in the supplementary material.

\subsection{Relational DSM Vectors}
\label{relational_dsm}

We start with an example of bulleted text. Without loss of generality, we split the entire document into multiple paragraphs ensuring that each paragraph at the most contains only one set of bulleted text. Bulleted text are more probable to be hyponyms  (Lists / Enumerations also included) as with the the instance below.

\textit{``X contains the following:}
\begin{itemize} 
\item \textit{X1}
\item \textit{X2''} 
\end{itemize}
Here, X1 and X2 are hyponyms of X. Note that hypernym-hyponym exactly corresponds to one relationship where $v_a$ denotes the former and $v_b$ denotes the latter. This can be easily generalized to other relationships and document structure rules. 
Given two words $x$ and $y$, their probability of their hierarchical (hypernym-hyponym in this case) relationship specific to a document feature $i$ (in this case bulleted list) can be stated as follows:
$$\rho^{k}(x,y) = \frac{\sum_{j=1}^{m^k}v_a^{j,k}(x) \cap v_b^{j,k}(y)}{\sum_{j=1}^{m^k}v_a^{j,k}(x)}.$$
Here, $j$ refers to an entity mention, which in this case is the presence of the word $x$ in paragraph $j$ that contains a document feature, for example, a bulleted list. 
Note that the suffix $k$ corresponds to the index of the document structure measure. Say for the three measures bulleted text, footnotes, and title, the indexes are respectively k = 1, 2, and 3. 
The notation for the vectors $v_a$, $v_b$, and $v_c$ follow similar suite. In case a paragraph $j$ does not contain the document feature, the corresponding entries are 0 for both the numerator and denominator sub-portions. In some cases a document feature may contain more than one occurrence of a hypernym / word. We merely consider the above expression to have an indicator function and do not pursue on the track of multiple occurrences. Usually, the vectors $v_a$ and $v_b$ are specified in an opposite sense, where say $v_a^{i,j}(x) = 1$, $v_b^{i,j} = 0$ and vice-versa. Say, if a word occurs at the text preceding the bullets, they are directed towards the indicator function in $v_a$ and if they occur within the sub-bullets, those are accounted towards the indicator function in $v_b$. However, it can also be true that some words can be present in both the preceding text or sub-bullets, leading to both $v_a$ and $v_b$ taking the value of 1. 
Generalizing the above to all possible document features, we have the following.
$$\rho(x,y) = \sum_{k=1}^{Kr} w^k f^k(n_x,n^{k,x}) \rho^{k}(x,y),$$
where $w^k$ refers to the weight assigned to each document feature (pre-set by the user depending on the application) and $f^k(.)$ denotes an importance function corresponding to the occurrence of the entities both in the presence of the context and overall (presence and absence included). More specifically, $n^{k,x}$ refers to the number of times the word $x$ has occurred in text preceding the document feature in the document and $n_x$ denotes the overall number of times the word $x$ has appeared in the document. 
For a complete set of rules and document features, please refer to the supplementary material.Alternatively, instead of a summation, $\rho$ can be kept as a vector and given by $\rho(x,y) = \left\{ \rho^k(x,y)\right\}_{k=1}^{Kr},$ where $Kr$ denotes the total number of relative features.

\subsection{Generating DSM vectors}

In this section we'll describe our information retrieval based approach to generate document structure measure vectors for each pair of entities in the Wikipeople and TACRED datasets. We use Wikipedia pages to produce the document structure, but any corpus of documents with a templated format like electronic health records, customer invoices, legislation, regulatory documents could be used to generate these DSM vectors. Additionally, as discussed in Section \ref{our_approach}, data sources like tweets, chat conversations, emails also display templated document structure and hence can be processed for generating DSM vectors using the method described in this section.

As discussed in Section \ref{datasets} and shown on Table \ref{tab:datasets}, Wikipeople and TACRED are relational datasets and unstructured datasets respectively, which we have converted into graph formats for this work. Hence in the case of Wikipeople, we're only concerned with Person to Person relations and in TACRED, only on relationships between people and organizations.

In order to generate document structure measure for the relationship between any of the above pairs, we begin by computing the frequency of the occurrence of pairs of entities in the corresponding Wikipedia pages of these entities. We index different parts of the Wikipedia pages in a search index and retrieve frequencies of occurrences using the surface forms of the entities (names) as queries.

We observe that an indexing system that leverages document structure and ontological information to enrich the raw documents, has higher performance than regular document indexing. We treat the search index as a black box (which is usually the case in many cloud based implementations) and make all our improvements to the documents being indexed. We leave further improvements to the search index as future work.

\begin{figure}[htb]
    \centering
    \includegraphics[width=\columnwidth]{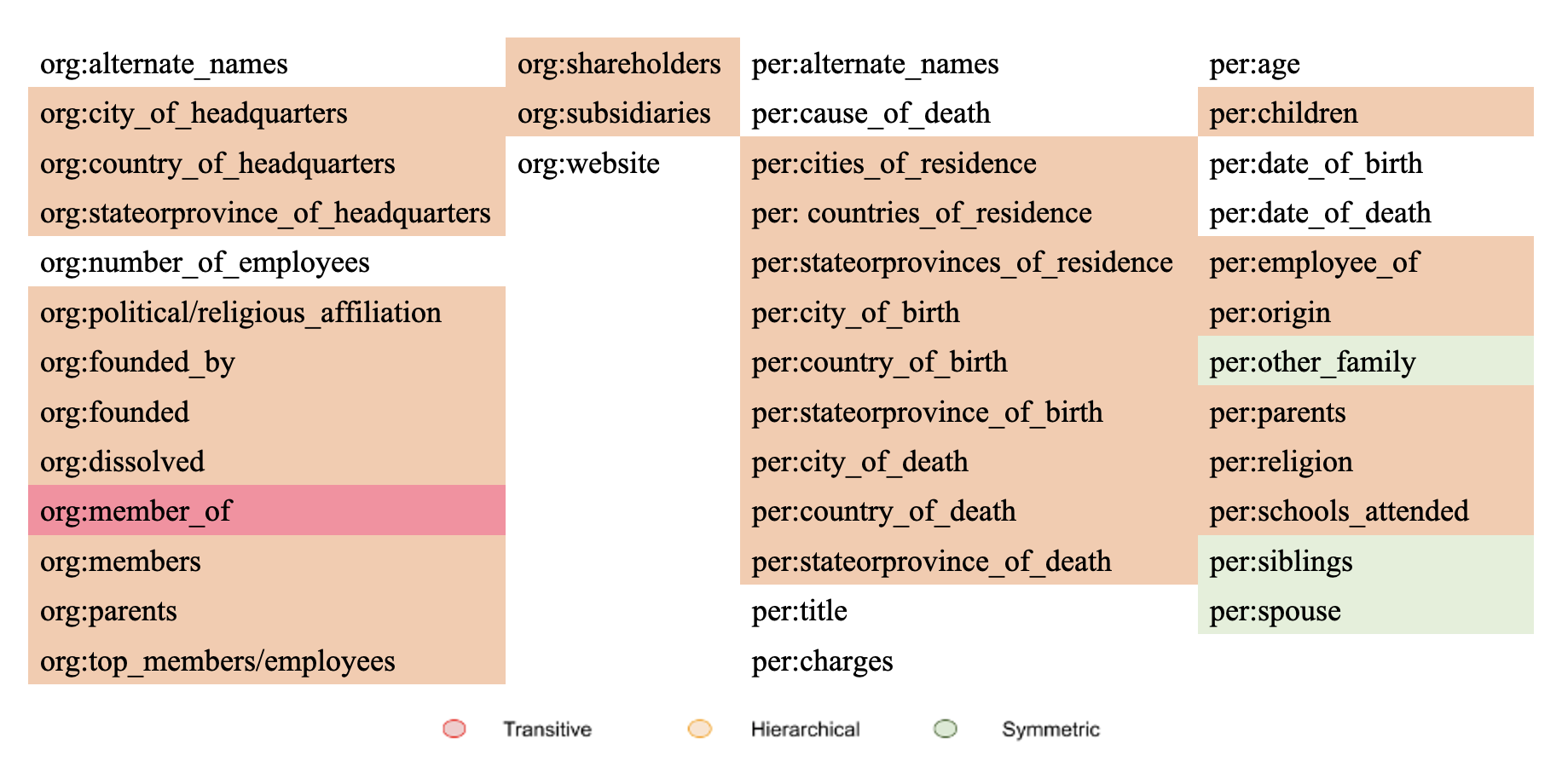}
    \caption{Relations in Tacred}
    \label{fig:tacred_relations}
\end{figure}

\begin{figure}[htb]
    \centering
    \includegraphics[width=\columnwidth]{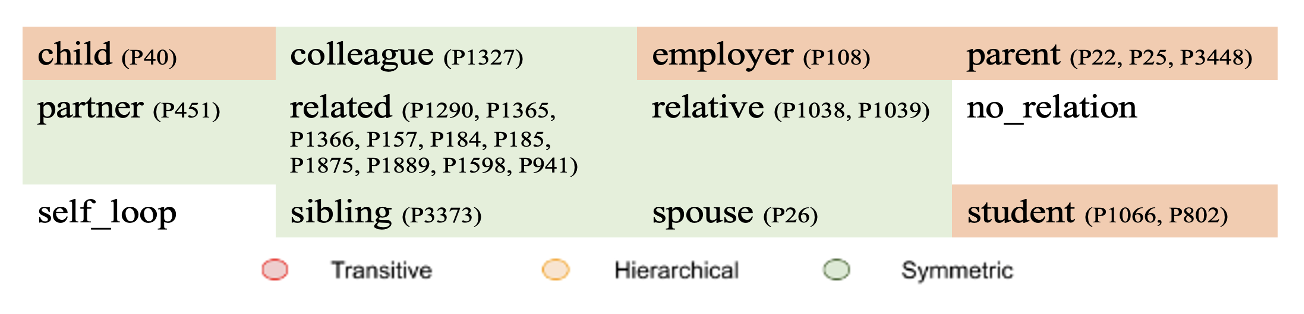}
    \caption{Relations in Wikipeople (with Wikidata property ids)}
    \label{fig:wikipeople_relations}
\end{figure}
\section{Experiments}

\subsection{Datasets}
\label{datasets}

We have created our datasets from the Wikipedia pages for the entities in Wikipeople and TACRED datasets. The original relations in the datasets are as shown in Figures \ref{fig:wikipeople_relations} and \ref{fig:tacred_relations}.

We first used NLTK\cite{loper2002nltk} to split the text into sentences and tokenize the sentences. We used the method described in~\cite{chiticariu2010systemt} to identify the span of entity mentions in Wikipedia pages. We call these scripts as Personal Data Annotators. This method requires creation of dictionaries each named after the entity type, and populated with entity mentions. We ran the Personal Data Annotators on these sentences, providing the bulk of the annotations that are reported in Table~\ref{tab:datasets}.

\begin{table}[!htp]
    \begin{center}
    \begin{tabular}{lrr}
        \hline
        \textbf{}   & \textbf{Wikipeople} & \textbf{Tacred} \\
        \hline
        \textbf{Documents}  & 16845 & 4142   \\
        \textbf{Nodes}  & 9760 & 11161 \\
        \textbf{Node Types} & 1 & 2 \\
        \textbf{Relation Types} & 13 & 22 \\
        \hline
    \end{tabular}
    \caption{Statistics on datasets adopted for relation type prediction in graphs}
    \label{tab:datasets}
    \end{center}
\end{table}

We consider the following document structures in this experiment. Title, introduction section, infobox, section (and subsection) headings and text. We could include other structures like category, in and out-links in the case of Wikipedia. We leave that for future work. To calculate the document structure measure for a relation, \{subject, relation, object\}, we use a weighted average of the frequency of their occurrences in the corpus. The weight of each document structure is manually assigned by us, but it could be learned from the corpus as well. We leave this also to future work.

This approach does not take the context of the entity mentions while assigning labels and hence the data is somewhat noisy. However, labels for name, email address, location, website do not suffer much from the lack of context and hence were annotated using this tool.

\subsection{Experimental setup}
We experiment on the R-GCN models proposed by~\cite{schlichtkrull2018modeling,shen2020taxoexpan}. These have been used in the cases of link prediction and hyponym based taxonomy expansion respectively. For training R-GCN and our improvements, we use a single V100 GPU with 16GB memory. We use the RGCN implementation in DGL \cite{wang2019dgl} and the elastic search instance from a cloud provider. 

\subsection{Link Type prediction with DSM}

Assuming that the network is defined appropriately, the 
hidden representation for each node i at (l+1)\textsuperscript{th} layer can be written as follows.
\begin{equation} \label{eq:rgcn}
\centering
h_{i}^{l+1} = \sigma \left(W\textsubscript{0}\textsuperscript{(l)}h\textsubscript{i}\textsuperscript{(l)} + \sum_{r \in R} \sum_{j \in N\textsubscript{i}\textsuperscript{r}} \frac{W\textsubscript{r}\textsuperscript{(l)}h\textsubscript{j}\textsuperscript{(l)}}{C\textsubscript{i, r}} \right).
\end{equation}

\subsection*{Baseline}

We use the heterograph version of R-GCN from DGL. One of the limitations of using the heterograph version is the requirement to add a self-loop for all the singleton nodes in the graph. As discussed in Section \ref{datasets}, we train all our R-GCN models by consuming the datasets as attributed graphs. In the case of Wikipeople, the relations are of the form (Person, relation\_type, Person), and in the case of TACRED, we have relations of the form (Person, relation\_type, Person), (Person, relation\_type, Org), (Org, relation\_type, Person) and (Org, relation\_type, Org). The baseline RGCN results seem comparable to the results on attributed multiplex heterogenous networks in \cite{cen2019gatne}, \cite{vannur2020data}.

\subsection*{Regularization}
We began by incorporating the DSM scores for each pair of relationship in the cross entropy loss function similar to any regularization parameter. As shown in Table \ref{tab:rgcn_results}, this did not yield any improvement in performance.

\subsection*{Hidden layer}
We then tried incorporating our DSM vectors by adding another hidden layer to the network. We pass the output of the first hidden layer in RGCN to this new layer and update the representation of entities in the (l+1)th layer with the DSM vectors. Recall that the hidden representation of entities in (l+1)th layer in R-GCN can be formulated as shown in Equation \ref{eq:rgcn}.

\noindent We updated the first hidden layer output for an entity by multiplying with the corresponding aggregate DSM score of all the edges incident on the entity. This method gave marginal improvement in some relationship types, but hurts the performance in general. We observe that using the aggregate scores from DSM vectors for all edges does not help to capture the correlation between the document structure and the edge. Hence we tried incorporating the DSM vectors in the message passing layer, which we describe in the next section.

\subsection*{Edge Weights}

Here, we describe a way to incorporate document structure measure in the node representation. We add a key embodiment to Equation \ref{eq:rgcn} in the form of 
$\rho$, corresponding to document structure measures. Note that we pre-compute $\rho$ based on samples from the training data. For the expression below, we use the general vector version of $\rho$, with individual $\rho^k$'s contributing towards the activation $\theta$. This can be replaced by the scalar $\rho^k$, where the activation can be changed accordingly. Note that the dimension of inputs and outputs corresponding to the layer can be re-defined accordingly with respect to the dimensions of x and y. 
\begin{equation} \label{eq:rho}
\centering
h\textsubscript{i}\textsuperscript{l+1} = \sigma \left( W\textsubscript{0}\textsuperscript{(l)}h\textsubscript{i}\textsuperscript{(l)} + \sum_{r \in R} \sum_{j \in N\textsubscript{i}\textsuperscript{r}} \frac{W\textsubscript{r}\textsuperscript{(l)}h\textsubscript{j}\textsuperscript{(l)}}{ C\textsubscript{i, r}}\right) + \theta (\rho^{i} h\textsubscript{i}\textsuperscript{(l)}).
\end{equation}


We implemented the above by updating the edge weights in the forward pass of RGCN. A similar approach has been adopted in \cite{qin2021relationaware} but the initial edge weights there were based on the node type and were updated with self-adversarial training. We instead incorporate the DSM vector during message passing by adding DSM scores to each corresponding edge weight. As discussed in the next section, this improves RGCN accuracy in both the datasets, with the Wikipeople increasing by 15 points.

\subsection{Results}

We report the performance of RGCN models as shown in Table \ref{tab:rgcn_results}. Our RGCN model with DSM edge weights performs about 15\% better than the baseline model. We also observe that using the DSM vectors for regularization and as a hidden layer do not seem to help and those models performed worse than the baseline.

\begin{table}[!htb]
    \begin{center}
    \begin{tabular}{lcc}
    \hline
    \textbf{Model} & \textbf{Wikipeople} & \textbf{TACRED}\\ \hline
    RGCN & 0.73 & 0.58 \\
    RGCN+DSM regularization & 0.71 & 0.57 \\
    RGCN+DSM hidden layer & 0.67 & 0.59 \\
    RGCN+DSM edge weights & \textbf{0.88} & \textbf{0.67} \\
    \hline
    \end{tabular}
    \end{center}
    \caption{Accuracy of RGCN models on the WikiPeople and TACRED Datasets}
    \label{tab:rgcn_results}
\end{table}

\begin{figure*}[!htb]
\begin{subfigure}{\textwidth}
    \centering
    \includegraphics[height=6cm,width=\columnwidth]{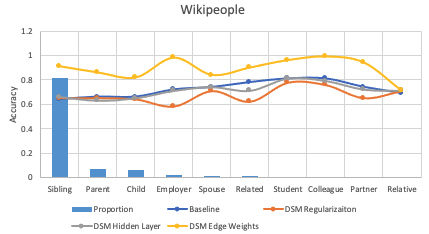}
    \caption{Wikipeople class distribution overlaid on accuracy of models}
    \label{fig:classwise_wikipeople}
\end{subfigure}
\begin{subfigure}{\textwidth}
    \centering
    \includegraphics[width=\columnwidth]{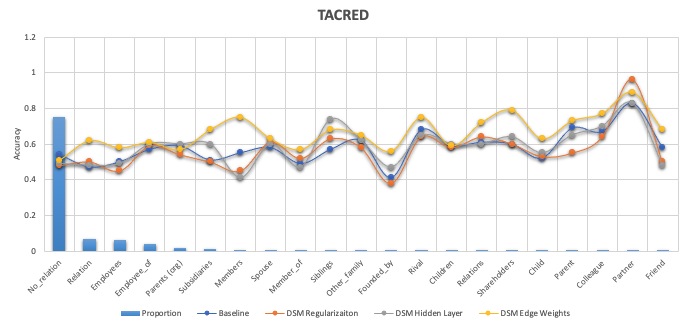}
    \caption{TACRED dataset distribution overlaid on accuracy of models}
    \label{fig:classwise_tacred}
\end{subfigure}
\caption{Class wise performance overlaid on accuracy of models}
\end{figure*}

In Figures \ref{fig:classwise_wikipeople} and \ref{fig:classwise_tacred}, we overlay the accuracy of the models on the class distribution. Sibling relationship type dominates the Wikipeople dataset and the use of document structure helps this type too. We believe that sibling relationship (which are symmetric) benefits from document structure because it gets the boost from the way we index sentences and associate the document and section title with the sentences. We also observe that no\_relation class which might rarely occur other than as text (and not in the infobox of a wikipedia page for example), does not get much improvement in accuracy.

We observe similar distribution in accuracy improvement for tacred, though the improvement is relatively less compared to Wikipeople. This seems to be because of the number of no\_relation examples in this dataset.

\section*{Conclusion}

In this work, we proposed using the document structure to improve ontological relation extraction from unstructured documents. In particular, we described a document structure measure (DSM) vector that can be incorporated while training a relational graph convolutional network (RGCN). We theoretically explained the need for such a measure in ontology population, and conducted experiments on different ways to incorporate the document structure measure in RGCNs. Our experiments show good improvement in RGCN performance while using our approach on the Wikipeople and TACRED datasets.

\clearpage
\bibliography{iclr2022_conference}
\bibliographystyle{iclr2022_conference}

\end{document}